\documentclass{article}

\PassOptionsToPackage{numbers, compress}{natbib}
\usepackage[final]{neurips_2019}

\usepackage[utf8]{inputenc} % allow utf-8 input
\usepackage[T1]{fontenc}    % use 8-bit T1 fonts
\usepackage{hyperref}       % hyperlinks
\usepackage{url}            % simple URL typesetting
\usepackage{booktabs}       % professional-quality tables
\usepackage{amsfonts}       % blackboard math symbols
\usepackage{nicefrac}       % compact symbols for 1/2, etc.
\usepackage{microtype}      % microtypography
\usepackage{textcomp}
\usepackage{amsmath}
\usepackage{amssymb}
\usepackage{color}
\usepackage{graphicx}
\usepackage{subfig}
\usepackage{wrapfig}
\usepackage{mathtools} 

\DeclareMathOperator*{\argmin}{arg\,min}
\DeclareMathOperator{\Tr}{tr}
\newcommand*\samethanks[1][\value{footnote}]{\footnotemark[#1]}

\setcitestyle{square}

\title{A Graph Autoencoder Approach to \\Causal Structure Learning}

\author{%
  Ignavier Ng\thanks{Work was done during an internship at Huawei Noah's Ark Lab.} \\
  University of Toronto \\
  \texttt{ignavierng@cs.toronto.edu} \\
  \And
  Shengyu Zhu \\
  Huawei Noah's Ark Lab \\
  \texttt{zhushengyu@huawei.com} \\
  \AND
  Zhitang Chen \\
  Huawei Noah's Ark Lab \\
  \texttt{chenzhitang2@huawei.com} \\
  \And
  Zhuangyan Fang\samethanks \\
  Peking University \\
  \texttt{fangzy\_math@pku.edu.cn}
}

\begin{document}

\maketitle

\begin{abstract}
  Causal structure learning has been a challenging task in the past decades and several mainstream approaches such as constraint- and score-based methods have been studied with  theoretical guarantees. Recently, a new approach has transformed the combinatorial structure learning problem into a continuous one and then solved it using gradient-based optimization methods. Following the recent state-of-the-arts, we propose a new gradient-based method to learn causal structures from observational data. The proposed method generalizes the recent gradient-based methods to a graph autoencoder framework that allows nonlinear structural equation models and is easily applicable to vector-valued variables. We demonstrate that on synthetic datasets, our proposed method outperforms other gradient-based methods significantly, especially on large causal graphs. We further investigate the scalability and efficiency of our method, and observe a near linear training time when scaling up the graph size.
\end{abstract}
\section{Introduction}

Causal structure learning has important applications in many areas such as genetics~\cite{koller2009probabilistic, PetJanSch17}, biology~\cite{Sachs2005CausalPN}, and economics~\cite{Pearl2019TheST}. 
An effective way to discover causal relations amongst variables is to conduct controlled experiments, which however is expensive and even ethically prohibited in certain scenarios. Learning causal structure from purely observational data has attracted much research attention in the past decades \cite{pearl2000causality,spirtes2000causation,shimizu2006linear,janzing2010causal}. 

Two major classes of structure learning methods from observational data are constraint- and score-based. Constraint-based methods, such as PC and FCI \cite{spirtes2000causation}, first use conditional independence tests to learn the skeleton of the underlying casual graph and then orient the edges based on a series of orientation rules \cite{meek1995causal, zhang2008completeness}. Score-based methods, like GES \cite{Chickering:2003:OSI:944919.944933}, assign a score to each causal graph according to some pre-defined score function \cite{heckerman1995learning, chickering1997efficient, bouckaert1993probabilistic} and then search over the space of causal directed acyclic graphs (DAGs) to find the one with the optimal score. However, finding the causal graph with optimal score is generally NP-hard, largely due to the combinatorial acyclicity constraint on the causal graph. Although the performance of several constraint- and score-based methods is  guaranteed with infinite data and suitable assumptions \cite{chickering1997efficient, spirtes2000causation}, the inferred graphs are usually less satisfactory in practice. More recently, \citet{zheng2018dags} proposed a smooth characterization on the acyclicity constraint, thus transforming the combinatorial optimization problem into a continuous one, provided with a proper loss function. Subsequent work \cite{yu2019dag} has also combined this approach with graph neural networks to model nonlinear causal relationships.

In this work, we propose a new gradient-based structure learning method which generalizes the recent gradient-based methods to a graph autoencoder framework that allows nonlinear structural equation models and vector-valued variables. We demonstrate that on synthetic datasets, our proposed method outperforms other gradient-based methods significantly, especially on relatively large causal graphs. We analyze the training time of our method and observe a near linear training time when scaling the graph size up to 100 nodes.

\section{Gradient-Based Causal Structure Learning} \label{sec:GAE}
This section introduces the recently developed gradient-based methods for causal structure learning. 

Let $\mathcal G$ be a causal DAG with vertices $ \{X_1, X_2,\cdots,X_d\}$, where $X_i, i=1,2,\ldots,d$ are vector-valued variables in $\mathbb R^l$. In this work, we focus on causal structure learning under additive noise models (ANMs) and adopt the same data generating procedure as in \cite{NIPS2008_3548}, that is, 
\begin{equation}
X_{i} \coloneqq f_i(X_{pa(i)}) + Z_i, i=1,2,\ldots,d, \nonumber
\end{equation}
where $X_{pa(i)}$ denotes the set of variables having a direct edge to $X_i$ in $\mathcal G$, $f_i: \mathbb{R}^{|X_{pa(i)}| \times l}\rightarrow \mathbb{R}^{l}$ is a vector-valued function, and $Z_i\in \mathbb{R}^{l}$ is an additive noise. We assume that $Z_i$'s are jointly independent, and also write $X\coloneqq[X_1, X_2,\cdots,X_d]$  and $Z\coloneqq[Z_1, Z_2,\cdots,Z_d]$. 

\citet{zheng2018dags} are probably the first to  transform the combinatorial optimization problem of score-based methods into a continuous one for linear structural equation model (SEM), which adopts the following data generation model
\begin{equation}
\label{eqn:linear_sem}
X_{i} = A_{i}^T X + Z_i, i=1,2,\ldots,d, \nonumber
\end{equation}
assuming both $X_i,Z_i\in\mathbb R$. Here $A_i\in\mathbb R^d$ is the coefficient vector and $A=[A_{1}, A_2,\cdots,A_d] \in \mathbb{R}^{d\times d}$ denotes the weighted adjacency matrix of the linear SEM. To ensure the acyclicity of $\mathcal G$, a smooth constraint on $A$ is proposed as
\[\Tr \left(e^{A \odot A}\right) - d = 0,\]
where $e^M$ denotes the matrix exponential of a matrix $M$ and $\odot$ denotes the Hadamard product. To learn the underlying causal structure, \citet{zheng2018dags} considers the regularized score function consisting of least squares loss and the $\ell_1$ norm, which, together with the above acyclicity constraint, leads to a continuous optimization problem:
\begin{equation}
\label{eqn:notears}
\begin{aligned}
\min_A \quad & \frac{1}{2 n}\sum_{j=1}^n\left\|X^{(j)}-A^T X^{(j)}\right\|_{F}^{2} + \lambda\|A\|_{1} \\
\textrm{subject to} \quad & \Tr \left(e^{A \odot A}\right) - d = 0,
\end{aligned}
\end{equation}
where $n$ is the sample size and $X^{(j)}$ denotes the $j$-th observed sample of $X$.

The above problem can then be solved by standard numeric optimization methods such as the augmented Lagrangian method~\cite{Bertsekas/99}. This approach is then named Non-combinatorial Optimization via Trace Exponential and Augmented lagRangian for
Structure learning (NOTEARS) by the authors. Though numeric optimization methods may not produce exact acyclic graphs, i.e., $\Tr(e^{A\odot A})-d $ can be very small (e.g., $10^{-8}$) but not exactly zero, post-processing like thresholding can be used to meet the hard acyclicity constraint. 

To generalize the above model to nonlinear cases, \citet{yu2019dag} proposed DAG-GNN, a generative model given below
\begin{align}
\label{eqn:DAGGNN}
  X=g_2\left(\left(I-A^{T}\right)^{-1}g_1(Z)\right),
\end{align}
where $g_1$ and $g_2$ are point-wise (nonlinear) functions. Causal structure learning is then formulated under the framework of variational autoencoders with multilayer perceptrons (MLPs) to model the causal relations, and the objective is to maximize the evidence lower bound under the acyclicity constraint. Notice that here $Z$ is regarded as the latent variable and its dimension can be selected to be lower than $d$, the number of variables.
\section{Graph Autoencoder for Causal Structure Learning}
We present an alternative generalization of NOTEARS to handle nonlinear causal relations. We consider the case with each variable $X_i$ being scalar-valued (i.e., $l=1$) in Section~\ref{sec:scaler} and then the the more general case ($l\geq 1$) in Section~\ref{sec:moregenearl}.
\subsection{Generalization of NOTEARS to Nonlinear Cases}
\label{sec:scaler}
We can rewrite problem~(\ref{eqn:notears}) as
\begin{align}
\label{eqn:notears_general}
\min_{A, \Theta} \quad & \frac{1}{2 n} \sum_{j=1}^{n}\left\|X^{(j)}-f(X^{(j)}, A)\right\|_{F}^{2} + \lambda\|A\|_{1} \\
\textrm{subject to} \quad & \Tr \left(e^{A \odot A}\right) - d = 0,\nonumber
\end{align}
where $f(X^{(j)}, A)$ denotes the data generating model w.r.t.~the $j$-th observed sample of $X$ and $\Theta$ denotes the parameters associated with $f$. For NOTEARS, we would have $f(X^{(j)}, A)=A^T X^{(j)}$ for linear SEM.

One way to extend $f(X^{(j)}, A)$ in Eq.~(\ref{eqn:notears_general}) to handle nonlinear relationships is to set
\begin{equation}
\label{eqn:first_generalization}
f(X^{(j)}, A) = A^T g_1(X^{(j)}),
\end{equation}
where $g_1:\mathbb{R}^l\to\mathbb{R}^l$ is a nonlinear function and in this work we use an MLP for this function. This generalization is a special case of causal additive model studied in \cite{Bhlmann2014CAMCA}. While it is possible to learn different MLPs for each variable, we use shared weights across all MLPs, similar to \cite{yu2019dag}.

To handle a more complex nonlinear model as in \cite{yu2019dag, Zhang:2009:IPC:1795114.1795190}, we consider to extend Eq.~(\ref{eqn:first_generalization}) to
\begin{equation}
\label{eqn:second_generalization}
f(X^{(j)}, A) = g_2\left (A^T g_1(X^{(j)})\right),
\end{equation}
where $g_1:\mathbb{R}^l \rightarrow \mathbb{R}^l$ and $g_2:\mathbb{R}^l \rightarrow \mathbb{R}^l$ are two MLPs with shared weights across all variables $X_i$.

% Note that how we construct $f(X^{(i)}, W)$ is similar to the network structure discussed in \cite{yu2019dag}. 
\subsection{Connection with Graph Autoencoder}
\label{sec:moregenearl}
\begin{figure}[htbp!]
\centering
        \includegraphics[width=0.75\textwidth]{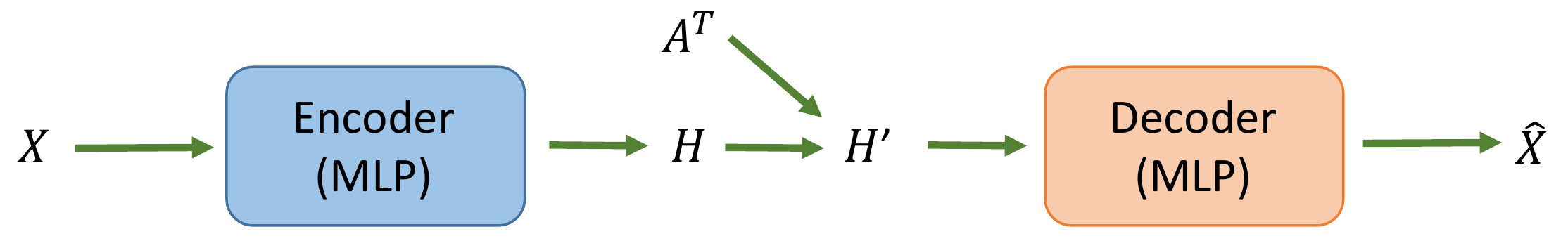}
    \caption{Model architecture for our proposed method.}
    \label{fig:gae_flow_chart}
\end{figure}

We now draw a connection of Eq.~(\ref{eqn:second_generalization}) with graph autoencoder (GAE), as outlined in Figure~\ref{fig:gae_flow_chart}. We can rewrite Eq.~(\ref{eqn:second_generalization}) as 
\begin{equation}
\begin{aligned}
\label{eqn:autoencoder}
H^{(j)} &= g_1(X^{(j)}), \\
H^{(j) \prime} &= A^T H^{(j)}, \\
f(X^{(j)}, A)  &= g_2(H^{(j)\prime}).
\end{aligned}
\end{equation}
Eq.~(\ref{eqn:autoencoder}), together with Eq.~(\ref{eqn:notears_general}), form a GAE (similar to \cite{DBLP:journals/corr/abs-1906-08745}) trained with reconstruction error where $g_1$ and $g_2$ are respectively the variable-wise encoder and decoder, and the message passing operation is applied at the latent representation $H^{(j)}$. In our case, the message passing operation is  a linear transformation $A^T H^{(j)}$, similar to the graph convolutional layer used in \cite{kipf2017semi}.

The GAE framework can easily include the vector-valued variable case. We  choose two variable-wise MLPs $g_1:\mathbb{R}^l \rightarrow \mathbb{R}^{l^{\prime}}$ and $g_2:\mathbb{R}^{l^{\prime}} \rightarrow \mathbb{R}^l$ where $l^{\prime}$ is a hyperparameter indicating the dimension of latent representation. One may set $l^{\prime} < l$ if $X$ is believed to have a lower intrinsic dimension. The final optimization problem is hence to minimize the reconstruction error of the GAE (with $\ell_1$ penalty):
\begin{equation}
\begin{aligned}
\label{eqn:final_optimization}
\min_{A, \Theta_{1}, \Theta_{2}} \quad & \frac{1}{2 n} \sum_{j=1}^{n}\left\|X^{(j)}-\hat{X}^{(j)}\right\|_{F}^{2} + \lambda\|A\|_{1} \\
\textrm{subject to} \quad & \Tr \left(e^{A \odot A}\right) - d  =0,
\end{aligned}
\end{equation}
where $\hat{X}^{(j)}=g_2\left(A^T g_1(X^{(j)})\right)$ is the reconstructed output  and $\Theta_1$ and $\Theta_2$ are the MLP weights associated with $g_1$ and $g_2$, respectively.

It is interesting to compare the proposed generalization Eq.~(\ref{eqn:second_generalization}) or Eq.~(\ref{eqn:autoencoder}) with Eq.~(\ref{eqn:DAGGNN}) used by DAG-GNN. The linear SEM can be written as $X=A^TX+Z$ with $X$ and $Z$ being the vectors concatenating all the observational and noise variables, respectively. We consider $Z$ as additive noises and Eq.~(\ref{eqn:second_generalization}) or Eq.~(\ref{eqn:autoencoder}) can be viewed as generalized causal relations taking $X_{pa(i)}$ as inputs for the $i$-th observational variable. In contrast, DAG-GNN further writes linear SEM as $X=(I-A^T)^{-1}Z$ and considers it as a generative model which takes random noises $Z$ as input. The experiments conducted show that our alternative generalization performs better, in both efficiency and efficacy, than DAG-GNN on similar datasets used by \cite{yu2019dag}.
\subsection{Augmented Lagrangian Method}
The optimization problem given in Eq.~(\ref{eqn:final_optimization}) can be solved using the augmented Lagrangian method. The augmented Lagrangian is given by
\[L_{\rho}(A, \Theta _{1}, \Theta _{2}, \alpha) = \frac{1}{2 n} \sum_{j=1}^{n}\left\|X^{(j)}-\hat{X}^{(j)}\right\|_{F}^{2} + \lambda\|A\|_{1}+\alpha h(A)+\frac{\rho}{2}|h(A)|^{2},\]
where $h(A):=\Tr(e^{A\odot A})-d$, $\alpha$ is the Lagrange multiplier, and $\rho>0$ is the penalty parameter. We then have the following update rules
\begin{align}
  A^{k+1}, \Theta_{1}^{k+1}, \Theta_{2}^{k+1} &= \argmin_{A,\Theta_1,\Theta_2} L_{\rho^{k}}(A, \Theta_{1}, \Theta_{2}, \alpha^k) \label{eqn:al_2}, \\
  \alpha^{k+1} &= \alpha^{k}+ \rho^{k}h(A^{k+1}) \label{eqn:al_3} \nonumber, \\
    \rho^{k+1} &=\begin{cases}
    \beta \rho^{k}, &~\text{if}~|h(A^{k+1})| \geq \gamma |h(A^{k})|,\\  
    \rho^{k}, &~\text{otherwise}, \nonumber
    \end{cases} 
\end{align}
where $\beta > 1 $ and $\gamma < 1$ are tuning hyperparameters. Problem~(\ref{eqn:al_2}) is first-order differentiable and we apply gradient descent method implemented in \texttt{Tensorflow} with Autograd and Adam optimizer \cite{Kingma2014AdamAM}.

\section{Experiments}\label{sec:experiments}

In this section, we conduct experiments on synthetic datasets to demonstrate the effectiveness of our proposed method. We compare our method against two recent gradient-based methods, NOTEARS \cite{zheng2018dags} and DAG-GNN \cite{yu2019dag}.

We use the same experimental setup as in \cite{yu2019dag}. In particular, we generate a random DAG using the Erd\H{o}s--R\'{e}nyi model with expected node degree $3$, then assign uniformly random edge weights to construct the weighted adjacency matrix $A$. We generate $X$ by sampling from ANM $X=f(A, X)+Z$ with some function $f$ elaborated soon. The noise $Z$ follows standard matrix normal. We report the structural Hamming distance (SHD) and true positive rate (TPR) for each of the method, averaged over four seeds. With sample size $n=3,000$, we conduct experiments on four different graph sizes $d\in\{10,20,50,100\}$. We consider scalar-valued variables ($l=1$) and vector-valued case ($l>1$) in Sections~\ref{sec:nonlinear_case} and ~\ref{sec:vector_case}, respectively.

For the proposed GAE, we use a 3-layer MLP with 16 ReLU units for both encoder and decoder. For NOTEARS \cite{zheng2018dags} and DAG-GNN \cite{yu2019dag}, we use the default hyperparameters found in the authors' code.

\subsection{Scalar-Valued Case}
\label{sec:nonlinear_case}

\begin{figure}
\centering
\subfloat[$X = A^T \cos(X + \textbf{1}) + Z$]
{
\includegraphics[width=0.92\textwidth]{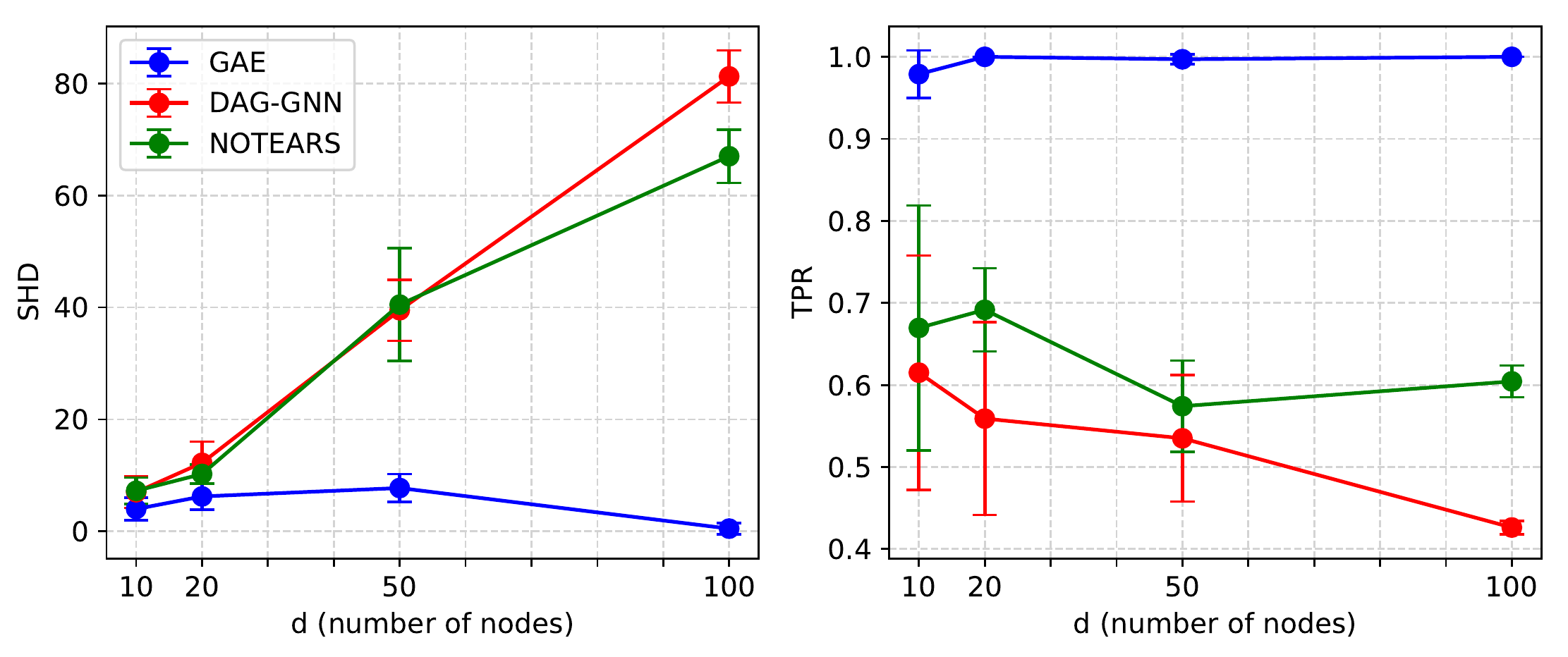}
\label{fig:nonlinear_1}
} \\
\vspace{-0.1in}
\subfloat[$X = 2\sin (A^T \cos(X + \textbf{1}) + 0.5 \cdot \textbf{1}) + (A^T \cos(X + \textbf{1}) + 0.5 \cdot \textbf{1}) + Z$]
{
\includegraphics[width=0.92\textwidth]{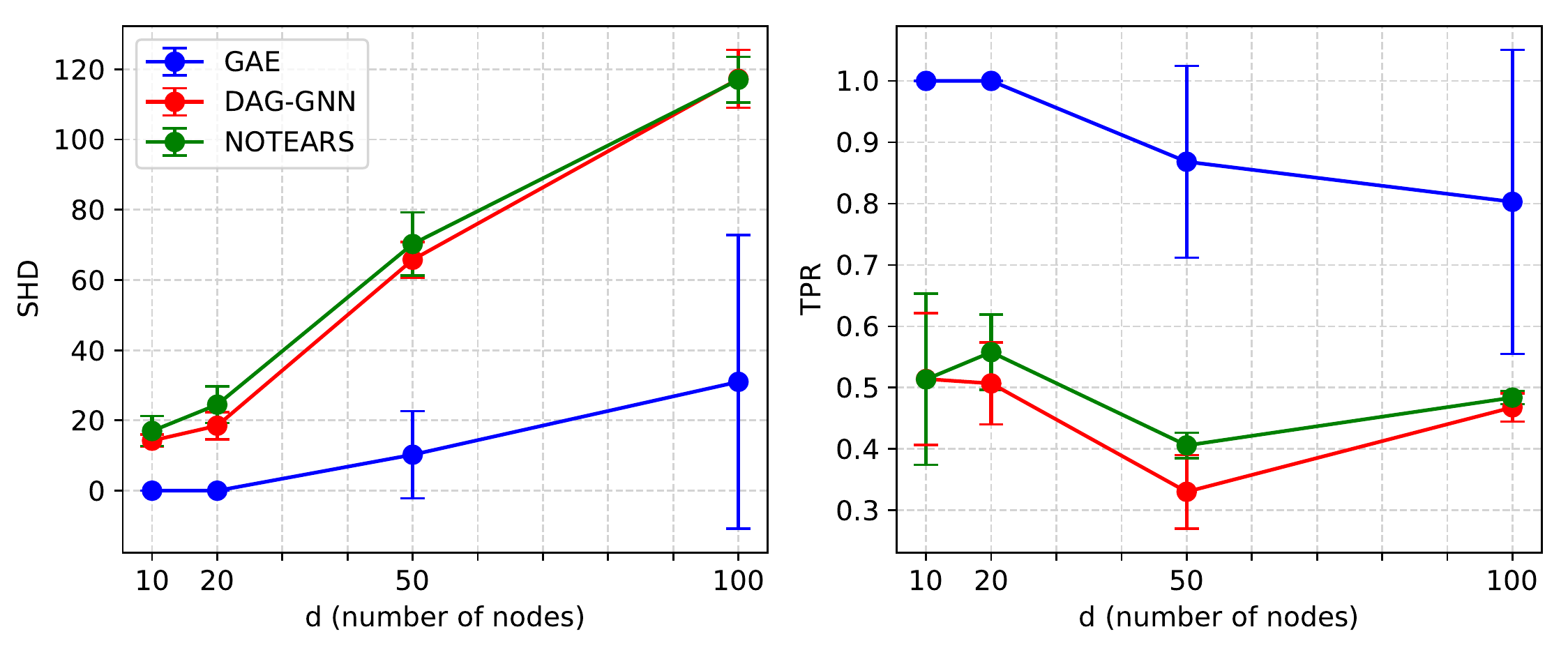}
\label{fig:nonlinear_3}
}
\caption{Experimental results on synthetic  datasets with scalar-valued variables.}
\label{fig:result_scalar_valued_dataset}
\end{figure}

Following the setup in \cite{yu2019dag}, our first dataset uses the data generating procedure below:
\begin{equation}
\label{eqn:nonlinear_1}
X = A^T \cos(X + \textbf{1}) + Z,
\end{equation}
which is a  generalized linear model. The results of SHD and TPR are reported in Figure~\ref{fig:nonlinear_1}. One can see that GAE outperforms NOTEARS and DAG-GNN, with SHD close to zero for graphs of 100 nodes. We also observe that NOTEARS has better performance than DAG-GNN when $d=100$, which indicates that DAG-GNN may not scale well on this dataset.

We next consider a more complicated data generating model, where the nonlinearity occurs after the linear combination of the variables:
\begin{equation}
\label{eqn:nonlinear_3}
X = 2\sin (A^T \cos(X + \textbf{1}) + 0.5 \cdot \textbf{1}) + (A^T \cos(X + \textbf{1}) + 0.5 \cdot \textbf{1}) + Z.
\end{equation}
As shown in Figure~\ref{fig:nonlinear_3}, our method has better performance than both NOTEARS and DAG-GNN in terms of SHD and TPR across all  graph sizes. We conjecture that with higher nonlinearity, the use of proposed GAE framework results in a much better performance than NOTEARS and DAG-GNN. For both nonlinear data generating procedure Eq.~(\ref{eqn:nonlinear_1}) and (\ref{eqn:nonlinear_3}), it is surprising that a linear model such as NOTEARS is on par with DAG-GNN or even has better performance in some cases. We hypothesize that the formulation of DAG-GNN results in the lack of causal interpretability on the adjacency matrix learned. 

% Figure with two sub-figures

\subsection{Vector-Valued Case}
\label{sec:vector_case}

To demonstrate the effectiveness of our method on vector-valued variables, we construct a dataset where each dimension of $X_i$ comes from a randomly scaled and perturbed sample. Given a graph adjacency matrix $A$, we first construct $\tilde{X}\in\mathbb R^{d\times 1}$ from the nonlinear SEM given in Eq.~(\ref{eqn:nonlinear_1}). We then generate for the $k$-th dimension $X^k=u^k\tilde{X}+v^k+z^k$, where $u^k$ and $v^k$ are scalars randomly generated from standard normal distribution and $z^k$ is a standard normal vector. The resulting sample is $X=[X^1,X^2,\cdots,X^l]\in\mathbb R^{d\times l}$. Our setup is similar to \cite{yu2019dag} but we use a nonlinear SEM in our experiment, which constitutes a more difficult vector-valued dataset.

\begin{figure}
\centering
        \includegraphics[width=0.92\textwidth]{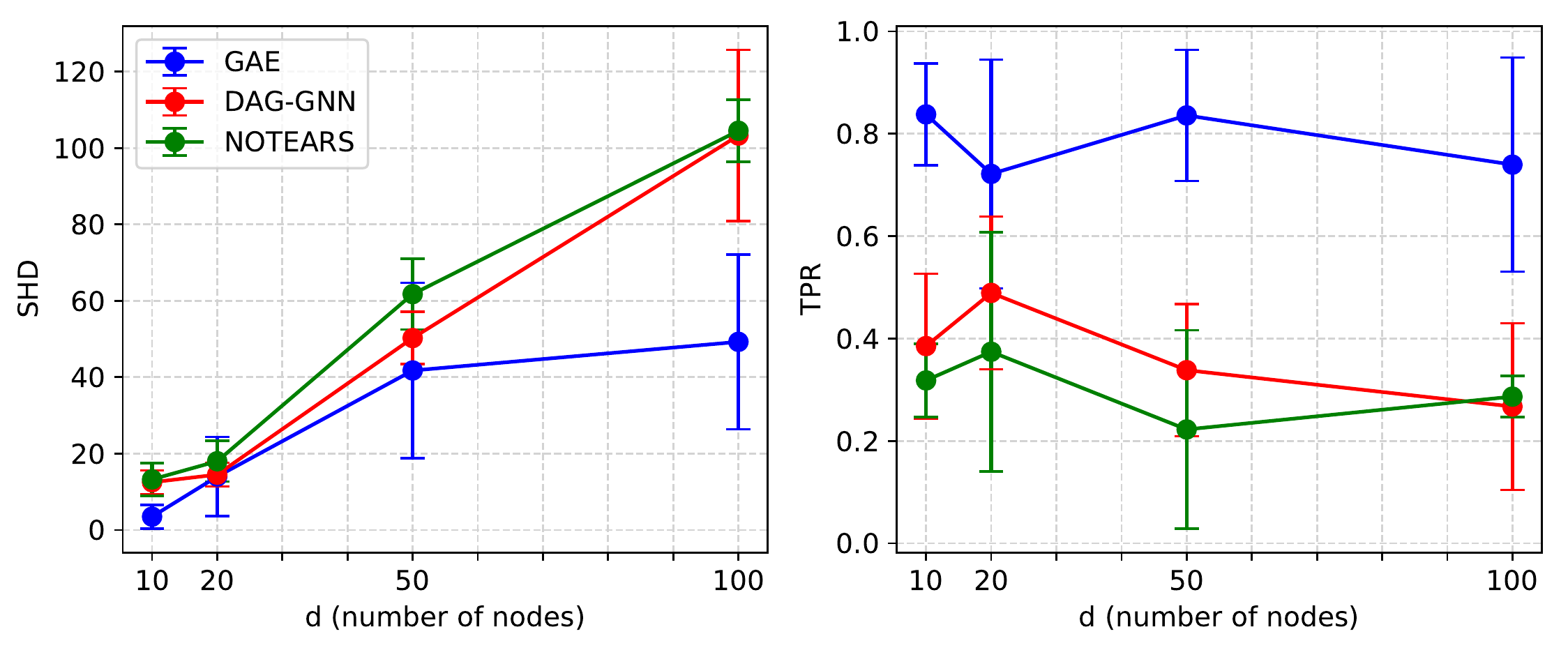}
    \caption{Experimental results on synthetic  datasets with vector-valued variables.}
    \label{fig:nonlinear_1_vector}
\end{figure}

In particular, we choose $l=5$ and the latent dimension of GAE $l^{\prime}=3$. The SHD and TPR are reported in Figure~\ref{fig:nonlinear_1_vector}, which shows that GAE has better SHD and TPR than NOTEARS and DAG-GNN especially when the graph size is large. It is also observed that DAG-GNN slightly outperforms NOTEARS for $d=20$ and $50$.
\subsection{Training Time}
\begin{figure}
\centering
        \includegraphics[width=0.45\textwidth]{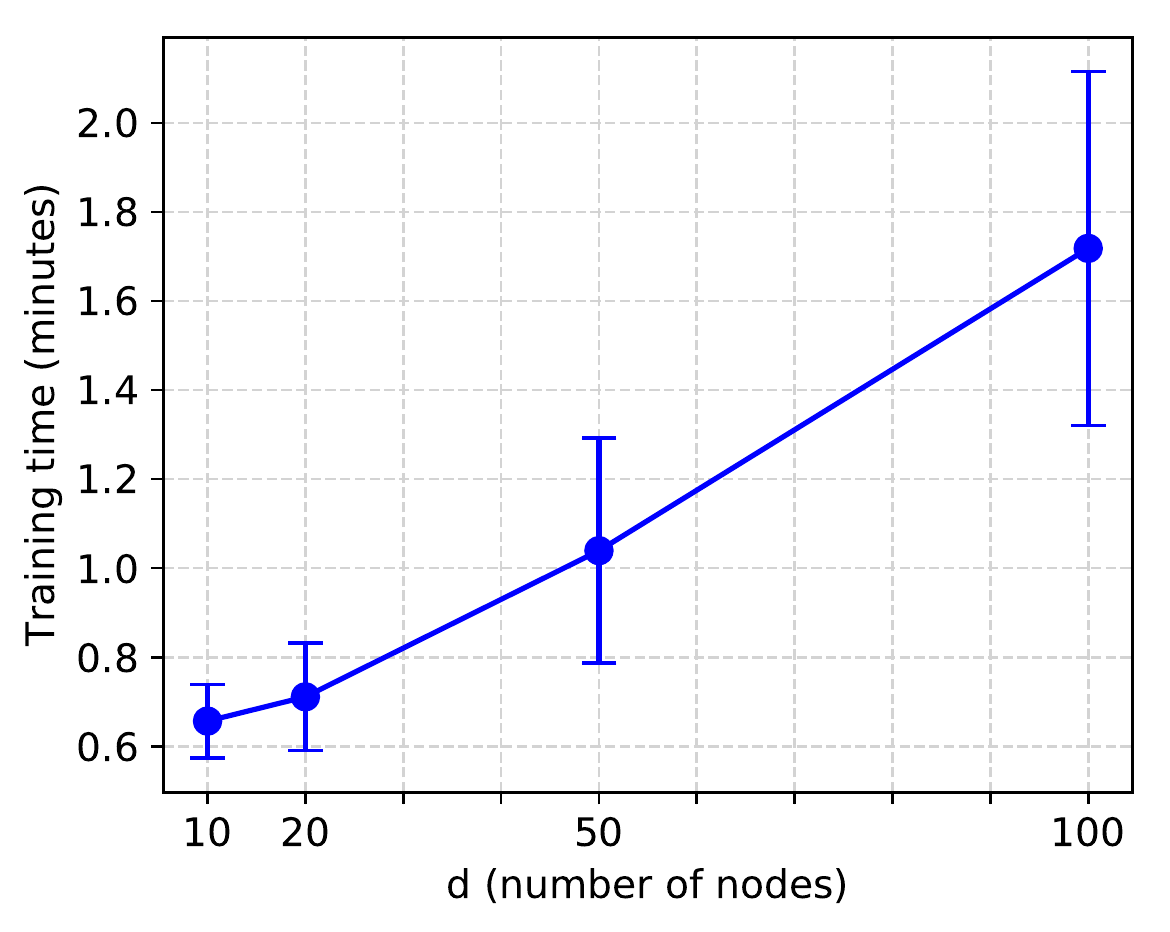}
    \caption{Average training time (minutes) of GAE across different graph sizes.}
    \label{fig:gae_training_time}
\end{figure}

Scalability is one of the important aspect in causal structure learning \cite{Sridhar2018ScalablePC}. To compare the efficiency and scalability of different gradient-based methods, we compute the average training time of GAE and DAG-GNN over all experiments conducted in Section \ref{sec:nonlinear_case} and \ref{sec:vector_case}. The experiments were computed on NVIDIA V100 Tensor Core GPU with 16GB of memory hosted on AWS GPU cloud instances. We do not include the average training time of NOTEARS as it uses only CPU instances.

As shown in Figure~\ref{fig:gae_training_time}, the average training time of GAE is less than 2 minutes even for graphs of 100 nodes. On the other hand, DAG-GNN training can be much more time-consuming: the training procedure takes $32.9\pm4.0$ and $77.0\pm18.0$ minutes for $d=10$ and $100$, respectively. We also observe that the training time of GAE seems to scale linearly when increasing the graph size to 100. The training time is fast as deep learning is known to be highly parallelizable on GPU \cite{Ben-Nun:2019:DPD:3359984.3320060}, which yields a promising direction in gradient-based methods for causal structure learning.

\section{Conclusion}

The formulation of continuous constrained approach by \cite{zheng2018dags} enables the application of gradient-based methods on causal structure learning. In this work, we propose a new gradient-based method that generalize the recent gradient-based methods to a GAE framework that allows nonlinear SEM and vector-valued datasets. On synthetic datasets, we demonstrate that our proposed method outperforms other state-of-the-art methods significantly especially on large causal graphs. We also investigate the scalability and efficiency of our method, and observe a near linear training time when scaling the graph size up to 100 nodes. Future works include benchmarking our proposed method on different graph types and real world datasets, as well as testing it on synthetic dataset with larger graph size (up to 500 nodes) to verify if the training time still scales linearly.

\bibliographystyle{plainnat}
\bibliography{ms}

\end{document}